\documentclass[pdflatex,sn-vancouver-num]{sn-jnl}

\usepackage{graphicx}%
\usepackage{multirow}%
\usepackage{amsmath,amssymb,amsfonts}%
\usepackage{amsthm}%
\usepackage{mathrsfs}%
\usepackage[title]{appendix}%
\usepackage{xcolor}%
\usepackage{textcomp}%
\usepackage{manyfoot}%
\usepackage{booktabs}%
\usepackage{algorithm}%
\usepackage{algorithmicx}%
\usepackage{algpseudocode}%
\usepackage{listings}%

\theoremstyle{thmstyleone}%
\theoremstyle{thmstyletwo}%

\theoremstyle{thmstylethree}%

\begin{document}

\title[Article Title]{Improving Credit Card Fraud Detection with an Optimized Explainable Boosting Machine}


\author*[1]{\fnm{Reza} \sur{E. Fazel}}\email{re.fazel22@gmail.com}

\author[2]{\fnm{Arash} \sur{Bakhtiary}}\email{bakhtiary.arash@gmail.com}

\author[3]{\fnm{Siavash} \sur{A. Bigdeli}}\email{sarbi@dtu.dk}

\affil*[1]{\orgdiv{Credit and Collection Department}, \orgname{EN Bank}, \city{Tehran}, \country{Iran}}

\affil[2]{\orgname{ReDi School}, \city{Copenhagen}, \country{Denmark}}

\affil[3]{\orgdiv{Department of Applied Mathematics and Computer Science}, \orgname{DTU}, \city{Copenhagen}, \country{Denmark}}


\abstract{Addressing class imbalance is a central challenge in credit card fraud detection, as it directly impacts predictive reliability in real-world financial systems. To overcome this, the study proposes an enhanced workflow based on the Explainable Boosting Machine (EBM)—a transparent, state-of-the-art implementation of the GA²M algorithm—optimized through systematic hyperparameter tuning, feature selection, and preprocessing refinement. Rather than relying on conventional sampling techniques that may introduce bias or cause information loss, the optimized EBM achieves an effective balance between accuracy and interpretability, enabling precise detection of fraudulent transactions while providing actionable insights into feature importance and interaction effects. Furthermore, the Taguchi method is employed to optimize both the sequence of data scalers and model hyperparameters, ensuring robust, reproducible, and systematically validated performance improvements. Experimental evaluation on benchmark credit card data yields an ROC-AUC of 0.983, surpassing prior EBM baselines (0.975) and outperforming Logistic Regression, Random Forest, XGBoost, and Decision Tree models. These results highlight the potential of interpretable machine learning and data-driven optimization for advancing trustworthy fraud analytics in financial systems.}




\maketitle

\section{Introduction}\label{sec1}

In today's digital economy, banks and credit card issuers face a critical responsibility: protecting customers from the growing threat of credit card fraud. As online transactions increase and fraud tactics become more sophisticated, the need to safeguard customers has become more urgent than ever. This protection is vital not only to uphold customer trust and financial security but also to ensure the integrity and stability of financial institutions.

Over the past decade, the incidence of credit card fraud has increased considerably, placing immense pressure on financial institutions\cite{dang_machine_2021}. The Nilson Report revealed that global card fraud losses reached \$33 billion in 2022, with the U.S. contributing roughly 40\% of these losses. Projections indicate that card fraud will remain a significant concern, potentially escalating to nearly \$400 billion globally by 2032\cite{castillo_why_2024}.

Machine learning (ML) methods have become essential tools in addressing this issue, enabling issuers to detect fraud, adapt to new fraud techniques, and enhance customer trust. However, the success of ML-based fraud detection depends heavily on the quality of the training data\cite{sanober_enhanced_2021,xue_dealing_2016}. Unlike traditional techniques, ML approaches can instantly sift through large volumes of transaction data to spot anomalies, continuously improving by learning from each incident. This approach increases precision and reduces false positives. An ML-based fraud prevention strategy involves data collection, training algorithms on recent fraud patterns, and ongoing model optimization. These steps protect customers and strengthen trust, establishing a secure financial environment. Credit card fraud datasets tend to be imbalanced, with a much larger number of genuine transactions compared to fraudulent ones, making it essential to use specialized techniques and avoid skewed model performance.

In financial contexts such as credit card fraud detection, having a model that is both high-performing and interpretable is critical. While many machine learning methods (e.g. boosted or bagged trees, SVMs, and deep neural networks) are effective for classification and regression tasks, their complexity makes the resulting models difficult for users to interpret\cite{lou_accurate_2013}.

To address these issues, we adopt a cutting-edge glass-box approach which manages data imbalance effectively without using resampling or anomaly detection methods.
Using Explainable Boosting Machines (EBM) \cite{nori_interpretml_2019}, we consistently achieve strong performance in the classification. A standout feature of the model is its ability to rank features by importance and impact on the target class, enabling selection of the most influential variables. This not only boosts model performance but also helps reduce complexity.

Our proposed approach includes a data preprocessing and hyperparameter tuning pipeline using the Taguchi method to further improve the accuracy of the classifier model.
In addition to identifying fraudulent transactions, our model can highlight the key factors that drive fraud detection, enabling card issuers to refine strategies and prune less significant features.

This paper leverages EBM to strike a balance between interpretability and high performance. Unlike black-box models, EBM constructs an additive model that displays the effects of individual features and pairwise interactions on predictions, aiding in feature selection. It often matches the performance of complex models like Random Forest and XGBoost, making it especially suitable for critical tasks such as fraud detection. In this work, we also compare EBM’s classification results with Logistic Regression, XGBoost, Random Forest, and Decision Tree models.

\section{Literature Review}\label{sec2}

In recent years, researchers have focused on AI-driven methods to improve credit card fraud detection, addressing the challenge of imbalanced datasets. These efforts aim to identify and compare the most effective models and techniques to enhance predictive accuracy. Fraud detection models examine examples of both fraudulent and legitimate transactions to determine whether a new transaction should be flagged as fraudulent \cite{alamri_hybrid_2024}. Supervised classification, which requires pre-labeled data, has been particularly effective. However, the scarcity of illicit transaction records compared to the total transaction volume poses a significant challenge due to class imbalance \cite{fiore_using_2019}. Many researchers have used advanced sampling techniques like SMOTE, whereas others rely purely on state-of-the-art models without explicitly adopting sampling methods.
\subsection{Sampling Methods and Their Variants}
Several studies have effectively applied Synthetic Minority Oversampling Technique (SMOTE) to address imbalanced datasets in credit card fraud detection, demonstrating its impact on improving model performance by generating synthetic samples for underrepresented classes \cite{dornadula_credit_2019, pundkar_credit_2023,rtayli_enhanced_2020, zhu_enhancing_2024}, while some have found under-sampling more effective to address data imbalance when using a specific model \cite{afriyie_supervised_2023}.

Hybrid resampling techniques have been effectively applied in recent years, yielding significant improvements in model performance \cite{abdul_salam_federated_2024, esenogho_neural_2022, alamri_hybrid_2024, sundarkumar_novel_2015}. Some recent research compares resampling methods. For instance \cite{gupta_unbalanced_2023} found that random oversampling and XGBoost yielded superior results, whereas \cite{praveen_mahesh_detection_2022} concluded that Random Forest performed best with under-sampled data. \cite{lucas_credit_2020} emphasized the importance of adaptable models for evolving fraud tactics. \cite{lokanan_predicting_2024} showed that ANN, combined with Benford’s law and SMOTE, improved Random Forest and Logistic Regression methods for money laundering detection. \cite{leevy_threshold_2023} compared threshold optimization with random under-sampling (RUS) and found threshold optimization more effective.
\subsection{Alternative Approaches}
Generative Adversarial Networks (GANs) are another effective method for handling data imbalance and have been extensively used in recent fraud detection research \cite{strelcenia_improving_2023, fiore_using_2019}. However, GANs require significant computational resources, and their performance can vary with architecture and training \cite{tanaka_data_2019}. Some studies report that GANs outperform traditional methods such as Adaptive Synthetic Sampling (ADASYN) and SMOTE \cite{tanaka_data_2019, ba_improving_2019}. Meanwhile, several researchers propose innovative techniques beyond existing resampling frameworks. \cite{li_hybrid_2021} used anomaly detection to remove outliers before training a non-linear classifier and employed a Dynamic Weighted Entropy and a Signal-to-Noise Ratio-based formula to handle class imbalance. \cite{carcillo_streaming_2018} applied Stochastic Semi-Supervised Learning and Active Learning strategies to enhance training diversity.

Autoencoders have also been integrated with various sampling strategies to address high-dimensional data challenges \cite{salekshahrezaee_effect_2023, abakarim_efficient_2018, zou_credit_2019}. Combining autoencoders with deep learning or machine learning models can outperform traditional methods \cite{pumsirirat_credit_2018, du_autoencoder_2023, tingfei_using_2020}.

Leveraging deep learning models to tackle the challenge of imbalanced datasets has been a focus of several studies in recent years and has acquired strong results \cite{mienye_deep_2023, najadat_credit_2020, gomez_end--end_2018, aziz_role_2023}.

There are other research studies that forego explicit resampling, often relying on tree-based approaches that inherently handle imbalance by focusing on the most informative splits.

A cost-sensitive decision tree, which focuses on high-value fraudulent transactions was developed by \cite{sahin_cost-sensitive_2013}. \cite{jain_performance_2020} assessed Decision Tree, Random Forest, and XGBoost for fraud detection, identifying XGBoost as the most effective. They addressed class imbalance via adjusted class weights and XGBoost’s scale\_pos\_weight, highlighting the importance of metrics beyond accuracy for robust evaluation. \cite{chhabra_roy_sustainable_2023} selected Random Forest for its ability to handle high-dimensional data and manage class imbalance via ensemble learning. In another study by \cite{carcillo_combining_2021} a hybrid approach was proposed where unsupervised methods like Isolation Forests, k-means, and Autoencoders to detect anomalies, and supervised methods like Random Forests, SVM, and Neural Networks were employed for classification using labeled data. \cite{leevy_comparative_2023} found that CatBoost outperformed other methods (Random Forest, XGBoost, Logistic Regression) for handling data imbalance. \cite{venkata_suryanarayana_machine_2018} examined various ML methods for credit card fraud detection, identifying Logistic Regression as the most effective for predicting fraud based on transaction features. \cite{islam_ensemble_2023} introduced the CCAD model, combining ensemble learning and meta-learning for anomaly detection. \cite{nami_cost-sensitive_2018} proposed a cost-sensitive fraud detection model that combines dynamic Random Forest and KNN, effectively handling class imbalance and improving accuracy.

Generalized additive models (GAMs) are considered the benchmark for interpretability when only univariate terms are included\cite{hastie_generalized_1990,lou_intelligible_2012}. Unfortunately, there is often a considerable gap in accuracy between the best standard GAMs and more complex models, such as tree ensembles\cite{lou_intelligible_2012}. \cite{lou_accurate_2013} introduced a model class known as GA²M that incorporates pairwise interactions while retaining intelligibility. \cite{nori_interpretml_2019} later developed an open-source python package that implementing an efficient GA²M-based algorithm called the Explainable Boosting Machine (EBM), which achieves accuracy comparable to state-of-the-art methods such as Random Forest and Boosted Trees but remains highly interpretable.

Although many recent studies apply Reinforcement Learning (RL) for HPO, we deliberately avoid RL for two key reasons. First, fraud detection datasets are typically highly imbalanced, as fraudulent events rarely occur, and offline RL often exhibits poor performance under such conditions \cite{jiang_offline_2023}. Second, RL is extremely sensitive to its own hyperparameter settings \cite{khadka_collaborative_2019,eimer_hyperparameters_2023,jomaa_hyp-rl_2019}, requiring additional hyperparameter tuning—a repetitive and time-consuming process that contradicts our efficiency goals.

In this work, we employ the Taguchi approach—a form of Design of Experiments—to address two main objectives: determining the optimal sequence of data scalers prior to model training, and performing hyperparameter optimization (HPO) for our model. The Taguchi method substantially reduces the number of experiments required, leading to significant savings in both computational cost and execution time.

\section{Dataset Description}\label{sec3}
European credit card fraud detection dataset from Kaggle is used in this research, comprising 284,807 transactions. The dataset consists of 30 features, and a 31st column (“Class”) indicates the response variable with two values: 0 for legitimate transactions and 1 for frauds. For privacy, features V1 through V28 have been transformed via Principal Component Analysis (PCA), but “Time” and “Amount” remain unmodified.

The dataset is highly imbalanced, with 284,315 samples in class 0 and 492 in class 1, resulting in a class ratio of 1:577. We do not employ any resampling approach due to concerns about losing large portions of majority-class data (in under-sampling) or generating potentially non-informative synthetic samples (in over-sampling). Instead, we center our efforts on optimizing an advanced Explainable Boosting Machine model to achieve high accuracy.

\section{Exploratory Data Analysis}
\subsection{Correlation analysis}
We examined various relationships—linear, monotonic, and complex— among the features using correlation matrices. The Pearson correlation matrix yielded values ranging from -0.53 to 0.4, indicating only weak to moderate relationships. To explore potential monotonic associations, we also examined Spearman’s and Kendall’s correlation coefficients. While Spearman and Kendall highlighted a moderate correlation between variables V21 and V22, this relationship was not captured by Pearson. According to \cite{van_den_heuvel_myths_2022}, Spearman and Kendall sometimes outperform Pearson in detecting linear associations, while Pearson can be more sensitive to certain nonlinear relationship patterns. The above observations merit further exploration, detailed in our Experimental Results section.
\subsubsection{Chatterjee’s Correlation Coefficient}
Pearson, Spearman, and Kendall are all strong tools for detecting linear or monotonic relationships; however, they do not detect non-monotonic interactions, even when the data are noise-free \cite{chatterjee_new_2021}. \cite{chatterjee_new_2021} introduced a correlation coefficient $\xi_{\mathbf{n}}$, and demonstrated that as the size of the data set (n) grows, this coefficient converges to a theoretical value $\xi(X,Y)$ that equals one if Y is a measurable function of X, and zero if X and Y are independent \cite{dalitz_simple_2024}. In this case we used XICOR package for R \cite{chatterjeexicor}, as researchers like \cite{hauke_comparison_2011} note that traditional correlation coefficients are limited to simpler relationships between two random variables.\\ 

\begin{figure*}[h]
\centering
\includegraphics[width=\linewidth]{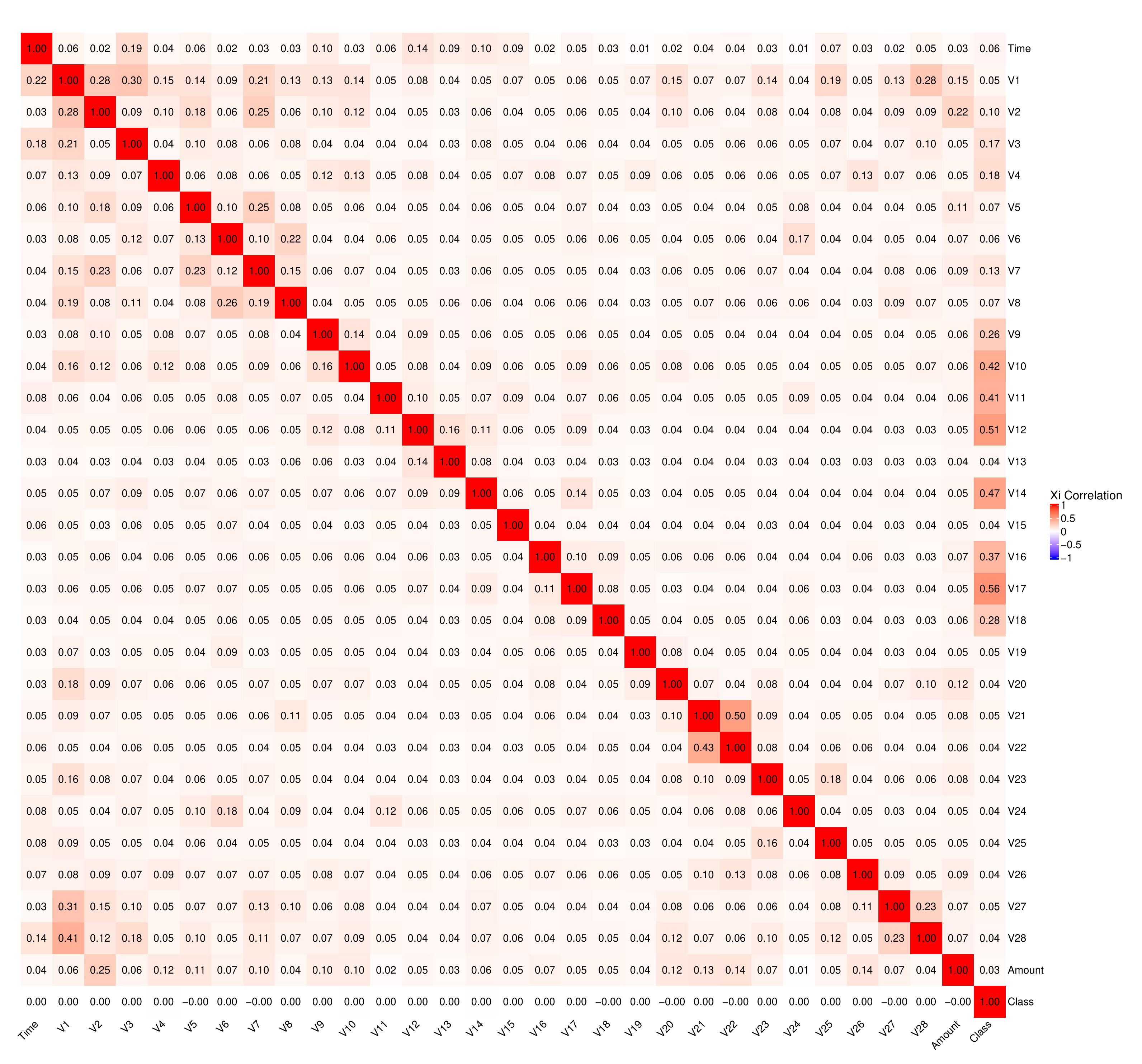}
\caption{Chatterjee’s Correlation Coefficient Heatmap}
\label{fig_1}
\end{figure*}

As clearly illustrated in Fig.~\ref{fig_1} which is plotted using the ComplexHeatmap package in R \cite{gu_complex_2022}, a moderate correlation between V21 and V22 is evident. Also there is a noteworthy correlation between V17 and Class which a discussion of these findings will be provided in the Experimental Results section. A Chatterjee’s correlation coefficient of 0.51 between V12 and Class and also 0.56 for V17 and the Class suggests a moderate degree of association between them. The Chatterjee correlation is a rank-based measure, making it robust to nonlinear and non-monotonic relationships. These numbers indicate the features likely provides meaningful information about the target.

\subsection{Multicollinearity Analysis}
If independent features are highly correlated, multicollinearity arises, and makes it difficult for the model to distinguish the effect of each feature on the dependent outcome.

To find multicollinearity, we use Variance Inflation Factor (VIF). It assesses how much the variance of an estimated regression coefficient is inflated when predictors are correlated \cite{akinwande_variance_2015}. The formula for the VIF is as follows:

\begin{equation}
\label{deqn_ex1a}
\text{VIF}_i = \frac{1}{1 - R_i^2}
\end{equation}

Where $R_i^2$ is the coefficient of determination obtained by regressing the $i^{\text{th}}$ predictor variable against all other predictor variables in the model. Table 1 shows the results of VIF analysis for all features in dataset.

\begin{table}[h]
\caption{VIF Analysis for Detecting Multicollinearity\label{tab:table1}}
\centering
\begin{tabular}{@{}lllll@{}}
\toprule
Feature & VIF & Standard\_Error & CI\_Lower & CI\_Upper\\
\midrule
Time &	2.340	& 0.003	& 2.335	& 2.345\\
V1	& 1.638	& 0.001	& 1.636	& 1.641\\
V2	& 3.901	& 0.005	& 3.890	& 3.911\\
V3	& 1.321	& 0.001	& 1.320	& 1.322\\
V4	& 1.172	& 0.000	& 1.172	& 1.173\\
V5	& 2.764	& 0.003	& 2.758	& 2.771\\
V6	& 1.529	& 0.001	& 1.527	& 1.531\\
V7	& 2.604	& 0.003	& 2.598	& 2.609\\
V8	& 1.099	& 0.000	& 1.098	& 1.099\\
V9	& 1.038	& 0.000	& 1.038	& 1.038\\
V10	& 1.209	& 0.000	& 1.208	& 1.210\\
V11	& 1.080	& 0.000	& 1.080	& 1.081\\
V12	& 1.154	& 0.000	& 1.154	& 1.155\\
V13	& 1.003	& 0.000	& 1.003	& 1.003\\
V14	& 1.220	& 0.000	& 1.219	& 1.220\\
V15	& 1.014	& 0.000	& 1.014	& 1.014\\
V16	& 1.081	& 0.000	& 1.081	& 1.081\\
V17	& 1.227	& 0.000	& 1.226	& 1.228\\
V18	& 1.034	& 0.000	& 1.034	& 1.034\\
V19	& 1.041	& 0.000	& 1.041	& 1.041\\
V20	& 2.234	& 0.002	& 2.229	& 2.238\\
V21	& 1.103	& 0.000	& 1.103	& 1.103\\
V22	& 1.082	& 0.000	& 1.082	& 1.083\\
V23	& 1.149	& 0.000	& 1.149	& 1.150\\
V24	& 1.001	& 0.000	& 1.001	& 1.001\\
V25	& 1.014	& 0.000	& 1.013	& 1.014\\
V26	& 1.001	& 0.000	& 1.001	& 1.001\\
V27	& 1.010	& 0.000	& 1.009	& 1.010\\
V28	& 1.002	& 0.000	& 1.002	& 1.002\\
Amount	& 11.508	& 0.020	& 11.469	& 11.547\\
Class	& 2.094	& 0.002	& 2.090	& 2.098\\

\botrule
\end{tabular}
\end{table}

There are varying perspectives on how to interpret VIF results and whether a universal rule can be applied across all contexts. The debate arises from differences in research fields, dataset characteristics, and specific analytical goals, leading to differing opinions on the applicability of general thresholds.

\cite{akinwande_variance_2015} stated that a VIF between 5 and 10 suggests a notable correlation that may create difficulties. When VIF surpasses 10, it typically signifies severe multicollinearity, leading to unreliable regression estimates and requiring corrective action. \cite{obrien_caution_2007} cautioned against relying solely on standard VIF thresholds (e.g., above 4 or 10) as the sole criterion for handling multicollinearity, such as by removing predictors. He focused on the importance of considering other factors, such as sample size, which can greatly influence the variability of regression coefficients and provide a more thorough understanding of the issue.

\cite{marcoulides_evaluation_2019} examine commonly used thresholds for VIF and Tolerance Index (TI), indicating that a VIF greater than 10 signifies serious multicollinearity, while values over 5 signals potential concerns requiring further analysis. They caution against strictly applying these thresholds, as their applicability depends on factors like sample size and study context. Additionally, they highlight that confidence intervals for VIF provide deeper insights, particularly when point estimates are close to the thresholds.

The VIF output reveals that most variables, such as V1, V3, and V4, have VIF values well below 5, indicating low to moderate multicollinearity, and therefore do not pose important issues. However, the “Amount” variable has a VIF of 11.51, which is notably high, suggesting significant multicollinearity. Additionally, the wide confidence interval for “Amount” (ranging from 11.469 to 11.547) indicates a level of instability in the VIF estimate. This suggests that “Amount” may be highly correlated with other variables in the model. The standard errors for most variables are small, indicating precise estimates of VIF, but the larger standard error for “Amount” further points to its instability and the need for further actions. Overall, most predictors do not show concern for multicollinearity; however, we employ feature scaling and transformation for the whole dataset,  which offer an effective way to mitigate the impact of multicollinearity, especially for “Amount” variable. In this study, techniques such as normalization, standardization, and other data scaling methods are applied to adjust the feature values. Additionally, tree-based models, known for their robustness to multicollinearity, are utilized. Specifically, the research employs the Explainable Boosting Machine, a type of tree-based model, to enhance interpretability and model performance.\\

\subsection{Causality Analysis}
In this study, the DoWhy package in Python was used to perform causal inference. Results revealed minimal causal effects for all features except “Amount,” which showed a remarkable causal relationship with multiple features. Removing “Amount” from the dataset led to a decrease in model precision, underscoring its importance in improving predictive accuracy. \\

\section{Data Transformation}
In this phase, we apply five data transformers —normalization, standardization, power transform, quantile transform, and robust scaler—and use the GridSearchCV to optimize their hyperparameters. These transformers are applied to five classification models: Logistic Regression, Decision Tree, Random Forest, XGBoost, and EBM. We then use GridSearchCV to optimize the hyperparameters for each transformation.

Table 2 illustrates the performance outcomes of various models after applying five distinct scalers optimized with their best hyperparameters. As observed, the EBM notably outperforms the other methods on all four performance metrics. The next step involves identifying the best sequence of scalers. Testing all permutations of scaler orders is computationally expensive, so we employ the Taguchi method to streamline this process. The sequence yielding the highest ROC-AUC score is adopted as the optimal configuration.

\begin{table}[h]
\caption{Comparison of model performance after applying five distinct scalers, optimized with best hyperparameters.
It is clearly visible that EBM consistently outperforms other models.\label{tab:table2}}
\centering
\begin{tabular*}{\textwidth}{@{}llp{4cm}llll@{}}
\toprule
Scaler & Model & Best Parameters & Precision & Recall & ROC\_AUC & F1 Score\\
\midrule
minmax & LR &  feature\_range: (-1 , 1)\  & 0.845 & 0.545 & 0.773 & 0.663\\
standard & LR & with\_mean: False; with\_std: False & 0.792 & 0.691 & 0.845 & 0.738\\
quantile & LR &  n\_quantiles: 1000;  output\_distribution: 'uniform'  & 0.794 & 0.736 & 0.868 & 0.764\\
robust & LR &  quantile\_range: (25.0 ,  75.0)  & 0.825 & 0.600 & 0.800 & 0.695\\
power & LR & method': 'yeo-johnson' & 0.823 & 0.591 & 0.795 & 0.688\\
minmax & RF &  feature\_range: (0 , 0.5)  & 0.976 & 0.727 & 0.864 & 0.833\\
standard & RF &  with\_mean: False; with\_std: False  & 0.975 & 0.718 & 0.859 & 0.827\\
quantile & RF &  n\_quantiles: 1000; output\_distribution: 'normal'  & 0.975 & 0.718 & 0.859 & 0.827\\
robust & RF &  quantile\_range: (25.0 ,  75.0)  & 0.975 & 0.718 & 0.859 & 0.827\\
power & RF & method: 'yeo-johnson' & 0.952 & 0.718 & 0.859 & 0.819\\
minmax & DT &  feature\_range: (-1 , 1)  & 0.670 & 0.682 & 0.841 & 0.676\\
standard & DT &  with\_mean: True;  with\_std: True  & 0.639 & 0.691 & 0.845 & 0.664\\
quantile & DT &  n\_quantiles: 1000; output\_distribution: 'normal'  & 0.672 & 0.709 & 0.854 & 0.690\\
robust & DT &  quantile\_range: (25.0 ,  75.0)  & 0.670 & 0.682 & 0.841 & 0.676\\
power & DT & method: 'yeo-johnson' & 0.667 & 0.673 & 0.836 & 0.670\\
minmax & XGB &  feature\_range: (0 ,  1)  & 0.976 & 0.736 & 0.868 & 0.839\\
standard & XGB &  with\_mean: True ;  with\_std: True  & 0.976 & 0.736 & 0.868 & 0.839\\
quantile & XGB &  n\_quantiles: 1500 ;  output\_distribution: 'normal'  & 0.976 & 0.736 & 0.868 & 0.839\\
robust & XGB &  quantile\_range: (25.0 ,  75.0)  & 0.976 & 0.736 & 0.868 & 0.839\\
power & XGB & method: 'yeo-johnson' & 0.976 & 0.736 & 0.868 & 0.839\\
minmax & EBM &  feature\_range': (0 ,  1)  & 0.988 & 0.755 & 0.877 & 0.856\\
standard & EBM &  with\_mean: True ;  with\_std: True  & 0.988 & 0.755 & 0.877 & 0.856\\
quantile & EBM &  n\_quantiles: 1000 ;  output\_distribution: 'normal'  & 0.976 & 0.755 & 0.877 & 0.851\\
robust & EBM &  quantile\_range: (25.0 ,  75.0)  & 0.988 & 0.755 & 0.877 & 0.856\\
power & EBM & method: 'yeo-johnson' & 0.988 & 0.755 & 0.877 & 0.856\\

\botrule
\end{tabular*}
\footnotetext{Note: LR = Logistic Regression; RF = Random Forest; DT = Decision Tree; XGB = XGBoost}
\end{table}

\section{Optimization Techniques}
\subsection{Taguchi Method for Scalers’ Order}
Developed by Genichi Taguchi, the Taguchi method systematically tests multiple factors with fewer experiments by using orthogonal arrays (OAs), which have a high degree of statistical confidence \cite{kaustav_sarkar_book_2020}. It is a systematic approach to designing experiments that helps optimize process or product performance while minimizing variability. It uses orthogonal arrays (OAs), which are specially arranged tables where each row represents an experimental run and each column represents a factor. These arrays allow balanced and independent testing of factor levels, reducing the total number of experiments compared to a full factorial design.

In practice, factors and their levels are selected, an appropriate OA is chosen, and experiments are conducted according to the array. The results are analyzed using the Signal-to-Noise (S/N) ratio, which highlights both performance and robustness. This process identifies the optimal factor settings efficiently, saving time and resources while ensuring reliable outcomes.

Rather than exhaustively testing all possible arrangements of the scalers, this approach chooses a reduced set of configurations that captures all influential factor levels. Since the goal of this research is to assess five different scalers and determine their optimal ordering for preprocessing data, we employ the standard L25 Taguchi OA. The L25 design tests 25 specific combinations of scalers instead of all 120 possible arrangements, significantly reducing computational time and complexity. We use the standard L25 OA for five scalers and five positions, modifying rows to skip redundant repeats by placing 0 for 'do nothing.' Through StratifiedKFold cross-validation, each row’s AUC is computed, and the best sequence is selected.

\subsection{Hyperparameter Tuning for Models}
Hyperparameters are crucial for controlling model complexity and preventing underfitting or overfitting. At this stage, we determine the optimal hyperparameters for each model by employing a Taguchi L9 orthogonal array (4 × 3). This Taguchi-based method allows us to identify the best combination of 4 different parameters, each with 3 different values, while minimizing computational costs. Using the L9 orthogonal array, we can efficiently test only 9 experiments to find the optimal settings, rather than testing all 81 possible combinations (since testing all status combinations would require evaluating $3^4=81$ different setups). The orthogonal array approach strategically selects a subset of these combinations to ensure comprehensive coverage of the parameter space, significantly reducing the number of experiments needed while still identifying the best configuration for model performance. Table 3 presents the optimal hyperparameters, focusing on the four most influential ones, each evaluated across three distinct levels for all models.

\begin{table}[h]
\caption{Optimal Hyperparameters for All models using the Taguchi method\label{tab:table3}}
\centering
\begin{tabular}{@{}lp{7.5cm}l@{}}
\toprule
Model	& Optimal Hyperparameters	& ROC-AUC\\
\midrule
Logistic Regression & C: 1, penalty: l2, class\_weight: \{0: 1, 1: 10\}, solver: saga & 0.981\\
Random Forest & max\_depth: 10, n\_estimators: 200, class\_weight: None, max\_features: sqrt & 0.974\\
Decision Tree & max\_depth: 5, min\_samples\_split: 50, min\_samples\_leaf: 20, class\_weight: \{0: 1, 1: 10\} & 0.926\\
XGBoost	& learning\_rate: 0.1, max\_depth: 3, scale\_pos\_weight: 10, subsample: 1.0 & 0.980\\
EBM & interactions: 20, max\_bins: 256, learning\_rate: 0.05, max\_rounds: 100 &	0.984\\

\botrule
\end{tabular}
\end{table}

For class imbalance, we applied model\-specific techniques. For instance, in XGBoost, we used the scale\_pos\_weight parameter to adjust for class distribution, and similar rebalancing strategies were adopted where available in other models.

To examine how sensitive the models are to the structure of the orthogonal array and the number of levels used, the analysis was extended using an L27 design with five hyperparameters at three levels each. Although this allowed for a broader exploration of the hyperparameter space, it led to only minimal changes in performance.

\section{Methodology}
The Explainable Boosting Machine (EBM) is an advanced machine learning model combining high interpretability with accuracy comparable to models like Random Forest and XGBoost.
EBM is inspired by the Generalized Additive Model (GAM):
\begin{equation}
\label{deqn_ex1a}
g(E[y]) = \beta_0 + \sum f_j (x_j),
\end{equation}
where $g$ represents the link function, which is applied to the expected value of response variable $y$ and $\beta_0$ is the intercept.
GAM is applicable to different settings such as regression or classification \cite{nori_interpretml_2019}.

~\cite{lou_accurate_2013} observe that two-dimensional interactions can be visualized as heatmaps of $f_{ij} (x_i,x_j)$ on the two-dimensional $x_i, x_j$-plane.
This approach, called Generalized Additive Models plus Interactions (GA2M), makes an interpretable model with one- and two-dimensional components, while also improves the overall performance. The model is formulated as:
\begin{equation}
\label{deqn_ex1a}
g(E[y])=\beta_0+\sum f_j (x_j )+ \sum f_{ij} (x_i ,x_j),
\end{equation}
This additive part with the pair-wise elements ensures both robust accuracy and high interpretability \cite{nori_interpretml_2019}.

We use the determined scaler sequences and hyperparameter configurations with best performance to train our initial EBM model.
Our next goal is to identify the most significant features that contribute to the target variable (model prediction) in our binary classification task.
We then retrain the model using only the top features, reducing complexity and computational efficiency by decreasing runtime.
Finally, we train other machine learning models on these selected features to compare their performance against EBM.
As we show in the results, this approach provides a better balance between interpretability and predictive accuracy.

\begin{figure*}[t]
\centering
\includegraphics[width=\textwidth]{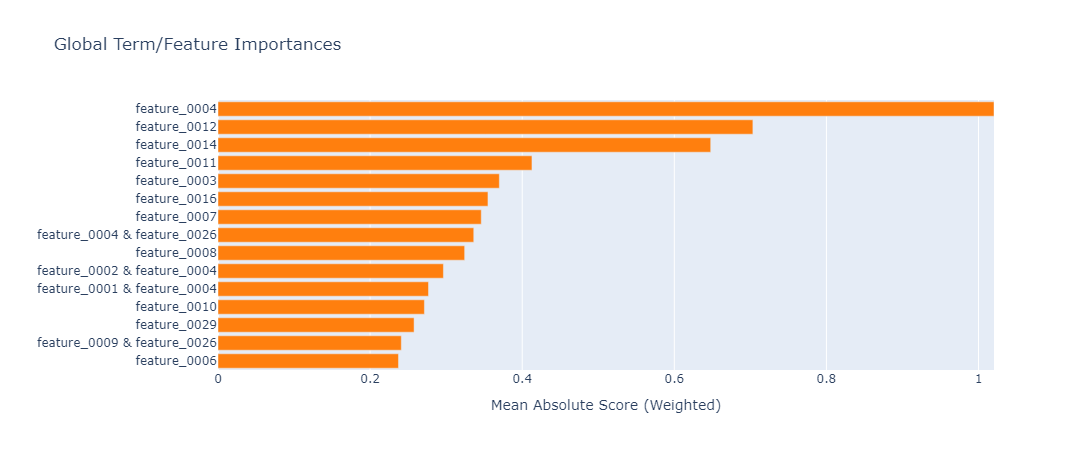}
\caption{Feature Importance and Pairwise Interactions }
\label{fig_2}
\end{figure*}

Fig.~\ref{fig_2} highlights the most significant individual features and pairwise interactions using EBM’s explain\_global() function, ranking them by their importance to the target class. EBM can automatically identify and incorporate meaningful two-dimensional interactions.

It learns each feature function $f_j$ using methods like bagging and gradient boosting. Boosting is limited to focus on one feature at a time in a round-robin fashion with a very low learning rate, ensuring that the order of features doesn’t affect results. This approach helps reduce co-linearity issues and learns the best function for each feature, clarifying how each contributes to the model’s predictions. Additionally, EBM can automatically detect and incorporate pairwise interactions of the formula (2) form which improves accuracy while keeping the model intelligible \cite{nori_interpretml_2019}.

\begin{figure*}[h]
\centering
\includegraphics[width=\textwidth]{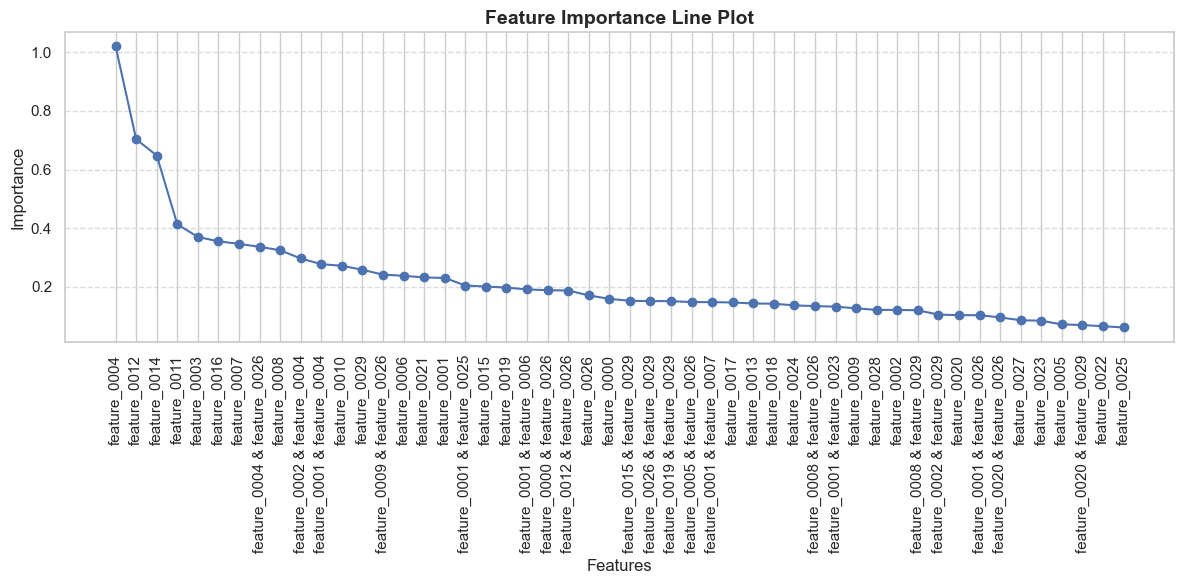}
\caption{Feature Importance and pairwise interaction of All Features from EBM’s Global Explanation}
\label{fig_3}
\end{figure*}

Fig.~\ref{fig_3} illustrates the ranked contributions of all individual features and pairwise interactions in class prediction.

EBM also furnishes local-level interpretability through the explain\_local() function, which shows how each feature contributes to an individual sample’s prediction.

\begin{figure*}[h]
\centering
\includegraphics[width=\textwidth]{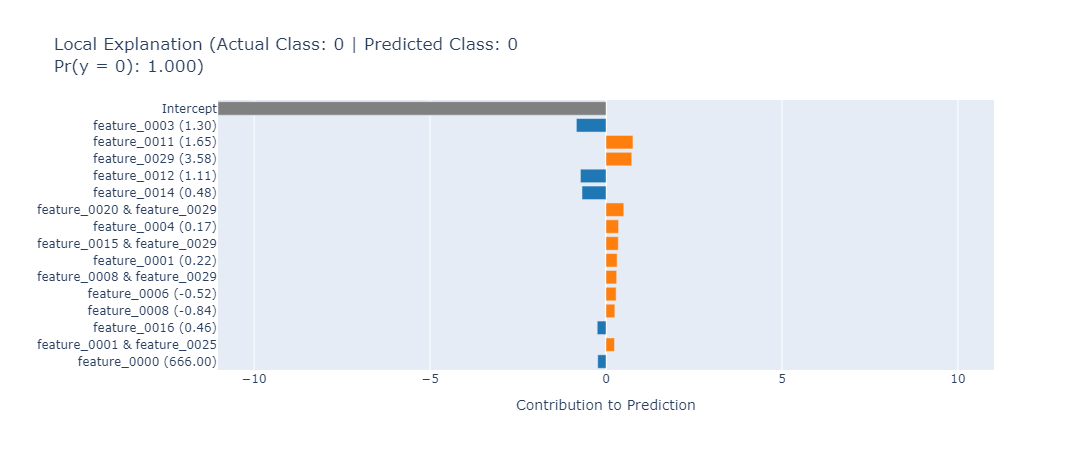}
\caption{Feature Contributions to a Class 0 Prediction}
\label{fig_4}
\end{figure*}

Fig.~\ref{fig_4} is a “local feature explanation” chart that shows the contributions of individual features to a machine learning model's prediction. Here’s what the picture communicates:  

The title indicates that the actual class is 0, the predicted class is 0, and the probability of Class 0 (Pr(y=0)) is 1. This means the model is almost 100\% confident in predicting the class as 0. In the bar chart explanation, the list of features is shown on the y-axis, where these features contributed to the final prediction. The intercept is displayed at the top as a baseline value for the prediction which represents the starting prediction before considering the feature contributions. The “blue bars” represent negative contributions (pulling the prediction towards class 0), while the “orange bars” represent positive contributions (pushing the prediction towards the opposing class, but not enough to change the outcome). The magnitude of the contributions is reflected by the length of the bars, with those extending further to the left or right indicating a greater impact on the prediction.  

In terms of interpretation, the most influential features that pushed the prediction towards class 0 include `feature\_0003`, `feature\_0012`, `feature\_0014` (which represent V3, V12 and V14), and others with substantial blue bars. Some features, like `feature\_0011` and `feature\_0029`, have smaller contributions shown as orange bars, which slightly counteract the dominant prediction but remain insufficient to shift the result. So the model predicts “Class 0” with extremely high confidence, and the key contributing features are those with significant negative (blue) contributions.

\begin{figure*}[h]
\centering
\includegraphics[width=\textwidth]{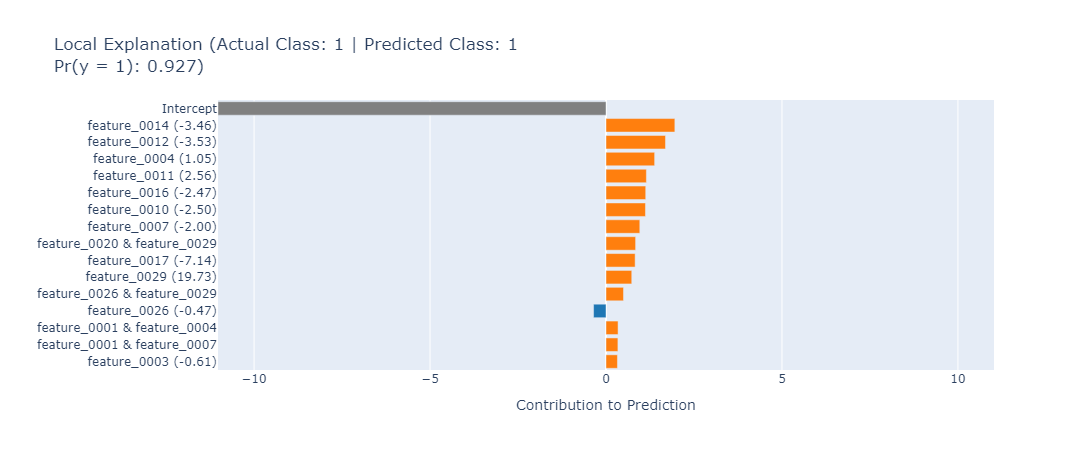}
\caption{Feature Contributions to a Class 1 Prediction}
\label{fig_5}
\end{figure*}

Fig.~\ref{fig_5} demonstrates how individual features contributed to the model's prediction for a specific sample with class 1. 
The prediction context shows that the actual class is 1, the predicted class is also 1, and the probability of class 1 is 0.927. This indicates that the model is highly confident, with a 92.7\% probability, that this sample belongs to class 1. The majority of the features contribute positively, as indicated by the orange bars. For instance, `feature\_0014` (wich represents V14) has the largest positive contribution, with a value of -3.46, making it a critical factor in driving the prediction toward class 1. Other features, such as `feature\_0012` and `feature\_0004`, also have significant positive contributions.  

Despite the presence of some negative contributions, the cumulative positive contributions dominate the prediction. As a result, the model predicts class 1 with high confidence. This chart effectively highlights the most influential features that drove the prediction toward class 1.

\section{Experimental Results and Discussion}
We trained the EBM for 3 to 30 top features using Stratified K-Fold cross-validation. To assess performance, we use Precision, Recall, F1 Score, and ROC-AUC, a key metric for imbalanced datasets \cite{brownlee_imbalanced_2020}. Table 4 shows the results.

\begin{table}[t]
\caption{Performance of EBM with Varying Number of Top Features\label{tab:table4}}
\centering
\begin{tabular}{@{}p{5cm}llll@{}}
\hline
Number of Top Features Contributing to the Prediction & Precision & Recall & ROC-AUC & F1 Score\\
\hline

3 & 0.802 & 0.706 & 0.975 & 0.750\\
4 & 0.825 & 0.687 & 0.976 & 0.745\\
… & … &	… & … & …\\
16 & 0.921 & 0.763 & 0.982 & 0.833\\
17 & 0.915 & 0.770 & 0.982 & 0.833\\
18 & 0.917 & 0.763 & {\bf 0.983} & 0.831\\
19 & 0.924 & 0.765 & 0.982 & 0.829\\
20 & 0.921 & 0.769 & 0.982 & 0.837\\
… & …	… & … &	…\\
30 & 0.922 & 0.755 & 0.981 & 0.835\\

\hline
\end{tabular}
\end{table}

The ROC-AUC score of 0.983 stands out as the highest achieved during testing of the proposed model and  surpasses the 0.975 reported by \cite{nori_interpretml_2019}, who introduced InterpretML as an open-source Python package and utilized EBM’s default parameters in their work, highlighting the effectiveness of our approach.

Notably, this score reaches 0.983 for the first time when utilizing the top 18 features. Reducing the number of features is crucial for minimizing model complexity and runtime, especially when working with large datasets. Table 5 shows the result of comparing the performance of EBM with other models by using top 18 features selected by EBM in training phase, which helps identify the most effective classifier.

\begin{table}[t]
\caption{Comparison of EBM and Other Models Using Top 18 Selected Features\label{tab:table5}}
\centering
\begin{tabular}{lllll}
\hline
Model & Precision & Recall & ROC-AUC & F1 Score\\
\hline

Logistic Regression & 0.765 & 0.806 & 0.979 & 0.780\\
Random Forest & 0.942 & 0.761 & 0.976 & 0.837\\
Decition Tree & 0.699 & 0.810 & 0.924 & 0.746\\
XGBoost & 0.806 & 0.822 & 0.977 & 0.814\\
EBM & 0.917 & 0.763 & 0.983 & 0.831\\

\hline
\end{tabular}
\end{table}

As discussed in previous sections, V21 and V22 (feature\_0021 and feature\_0022) exhibit a moderate degree of correlation that warrants further investigation. In our case, only V21 appears in the top 18 features and thus remains in the final model. Systematic removal did not yield further performance gains. Additionally, we observe that V12 is among the top 18 features selected by the EBM model, indicating its relevance. In contrast, V17, ranked 22nd by the EBM and excluded from the top 18 training features, does not contribute to improving the model’s performance.

\section{Addressing Overfitting and Model Robustness}
To assess overfitting, the cross\_validate function (with return\_train\_score=True) was used. The model’s average training score was 0.99858, and the average test score was 0.98185—an acceptable train-test gap of 0.01673 (1.67\%). Because this gap is much smaller than a set threshold of 0.1, we conclude that the model generalizes well. The negligible gap and high test performance confirm the absence of notable overfitting.

\section{Conclusion and Future Work}
This study highlights the effectiveness of EBM-driven feature selection in optimizing credit card fraud detection. By leveraging the model’s interpretability, we identified the top 18 most influential features, achieving an ROC-AUC score of 0.983 eliminating the need for all features, which would otherwise increase computational cost and model complexity. This eliminated the need for all 30 features and thereby reduced computational costs and complexity. An EBM trained on these selected features outperformed alternative models on the same feature set, reinforcing EBM’s accuracy advantage.

Additionally, we implemented an optimized preprocessing and hyperparameter tuning strategy leveraging the Taguchi method, which demonstrated superior efficiency and effectiveness compared to exhaustive techniques such as GridSearchCV. By systematically optimizing the sequence of preprocessing steps—including scaler selection—alongside hyperparameter configurations and feature subset selection, the Explainable Boosting Machine (EBM) achieved notable gains in predictive performance. In fact, EBM consistently outperformed several well-established models, including Logistic Regression, Random Forest, XGBoost, and Decision Tree, across key evaluation metrics.

Due to the combinatorial nature of hyperparameter tuning, it is impractical to exhaustively evaluate all possible orthogonal arrays or to test an extensive range of parameter levels within the scope of a single study. We encourage future research to explore alternative OA configurations, potentially involving more parameters or additional levels, to assess whether further performance improvements can be achieved.

To ensure robustness, we evaluated overfitting with cross-validation, revealing a minimal train-test gap (1.67\%), indicating strong generalization capabilities. Beyond its immediate contributions, this research opens new directions in feature selection and model optimization for fraud detection and other imbalanced classification problems. The interpretability of EBM can be further assessed in various datasets and domains, and the Taguchi method may find broader application in machine learning workflows.

While EBMs do not support online learning, future work could explore adapting them for real-time fraud detection through periodic mini-batch retraining. This approach allows efficient integration of new data while maintaining historical patterns, supporting both stability and adaptability. Preliminary experiments with such incremental strategies may offer valuable insights for further optimization. In addition, state-of-the-art RL models can adapt to evolving fraud tactics as feature importance shifts, but their exploration risks and high computational costs limit practicality.

Overall, our results establish a new benchmark in credit card fraud detection, introducing a highly effective approach to significantly enhance predictive accuracy. By reaching an ROC-AUC of 0.983, we exceed previous studies and illustrate the transformative potential of explainable AI (XAI) in financial fraud detection, offering a powerful and transparent solution for real-world applications.

\section*{Acknowledgements}
This research did not receive any specific grant from funding agencies in the public, commercial, or not-for-profit sectors.

\bibliography{references}

\end{document}